\documentclass[lettersize,journal]{IEEEtran}
\usepackage{amsmath,amsfonts}
\usepackage{amsmath}
\usepackage{amsthm}
\usepackage{amssymb}
\usepackage{algorithmic}
\usepackage{algorithm}
\usepackage{array}
\usepackage[caption=false,font=normalsize,labelfont=sf,textfont=sf]{subfig}
\usepackage{textcomp}
\usepackage{stfloats}
\usepackage{url}
\usepackage{verbatim}
\usepackage{graphicx}
\graphicspath{{./figures/}} 
\usepackage{cite}
\usepackage{hyperref}
\newtheorem{theorem}{Theorem}
\newtheorem{definition}{Definition}

\begin{document}

\title{Differentially Private Vertical Federated Learning}

\author{Thilina Ranbaduge and Ming Ding~\IEEEmembership{Senior Member,~IEEE,}
\thanks{T.Ranbaduge and M.Ding are with Data61, CSIRO, Australia
(e-mail: {Thilina.Ranbaduge; Ming.Ding}@data61.csiro.au)}}

\markboth{Journal of IEEE}%
{Shell \MakeLowercase{\textit{et al.}}: A Sample Article Using IEEEtran.cls for IEEE Journals}


\maketitle

\begin{abstract}
A successful machine learning (ML) algorithm often relies on a 
large amount of high-quality data to train well-performed models. 
Supervised learning approaches, such as deep learning techniques, 
generate high-quality ML functions for real-life applications, 
however with large costs and human efforts to label training data. 
Recent advancements in federated learning (FL) allow multiple data
owners/organisations to collaboratively train a machine learning 
model without sharing raw data. In this light, 
vertical FL allows organisations to build a global model when 
the participating organisations have vertically partitioned data. 
Further, in the vertical FL setting the participating organisation 
generally requires fewer resources compared to sharing data directly, 
enabling lightweight and scalable distributed training solutions.
However, privacy protection in vertical FL is challenging due to 
the communication of intermediate outputs and the gradients of model update. 
This invites adversary entities to infer other organisations’ underlying data. 
Thus, in this paper, we aim to explore how to protect the privacy of 
individual organisation data in a differential privacy (DP) setting. 
We run experiments with different real-world datasets and DP budgets. 
Our experimental results show that a trade-off point needs to be 
found to achieve a balance between the vertical FL performance and 
privacy protection in terms of the amount of perturbation noise. 



\end{abstract}

\begin{IEEEkeywords}
Differential privacy, clipping, stochastic gradient descent.
\end{IEEEkeywords}

\section{Introduction}

Artificial intelligence (AI) has gained significant attention because 
of the achievements of machine learning (ML) and deep learning algorithms  
that rapidly accelerate research and transform data processing practices 
in diverse business sectors, including health, agriculture, cybersecurity,
and advanced manufacturing~\cite{bonawitz2017,gong2020,mothukuri2021}.
Training in heterogeneous and potentially massive networks introduces 
novel challenges that require fundamental innovations in large-scale 
machine learning, distributed optimisation, and privacy-preserving data
analysis~\cite{aledhari2020}.

Federated learning (FL)~\cite{mcmahan2017} is a new learning paradigm that 
aims to build a joint ML model based on the data located at multiple 
sites or owned by different participants. In the model training, 
information such as gradients are exchanged between participants, but not 
the raw data. The exchanged information does not reveal any protected 
private portion of the data belonging to any party. 

\begin{figure}[!t]
\centering
\includegraphics[width=0.48\textwidth]{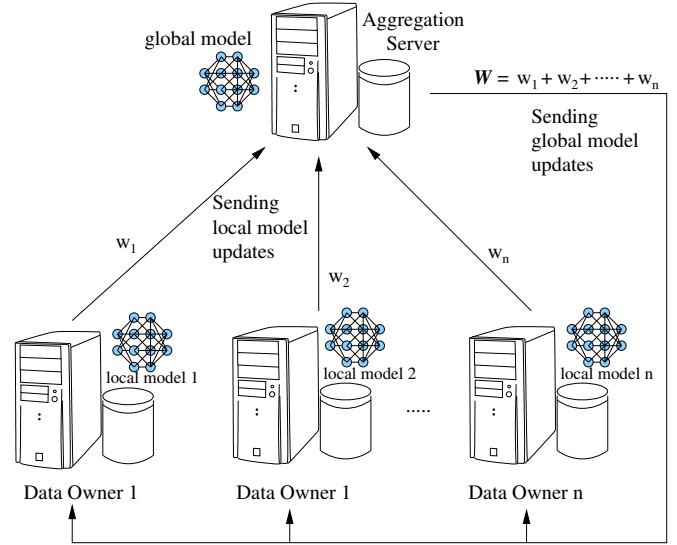}
\caption{An example federated learning architecture. In order to guarantee
the privacy of the data, federated learning only permits all the data owners
to exchange the model gradient with the aggregation server. During this 
process, each data owner trains its own model with the local data, then 
uploads the local model to the aggregation server. After aggregating all
the received models, the server returns the new global model to each 
data owner.}
\label{fig:fl-overview}
\end{figure}

Figure~\ref{fig:fl-overview} illustrates the FL process. At first each data 
owner will receive a generic global model from the aggregation server. 
Once the initial model is received, each data owner will conduct local
training on this model with their data separately and then upload the
related gradient information (local model updates) to the aggregation 
server. The aggregation server then averages the updates sent by data 
owners into the global model and then updates the global model to replace 
each user’s local model. The above steps repeat until the global model
achieves the required performance or the training reaches the maximum
iteration number.

In federated learning, privacy protection has become a major 
concern~\cite{aledhari2020,nguyen2021}. Althought federated learning protects
the private data on each device by exchanging model gradients with the server, 
instead of raw data, the model communication during the entire training
process can still leak sensitive information to a third party, e.g., the 
reverse engineering of models~\cite{zihao2022}. Although there are some
methods to improve the privacy of data recently, these methods tend to 
increase the computational burden of the federated network~\cite{nguyen2021}. 
In order to further protect the security of private data, we need to find
new methods to prevent private data from leakage during FL model 
transmission~\cite{li2021,wei2020}.

Federated learning can be categorised into three groups according to 
the distribution of data, i.e., horizontal federated learning, vertical 
federated learning, and federated transfer learning~\cite{li2020}. 
Horizontal federated learning is suitable in the case that the user features
of the two datasets overlap a lot, but the users overlap little. 
Vertical federated learning is available in the case that the user features
of the two datasets overlap little, but the users overlap a lot. 
In the case that the users and user features of the two datasets both 
rarely overlap, we can use transfer learning to overcome the lack of 
training data~\cite{liu2020}. 

\begin{figure}[!t]
\centering
\includegraphics[width=0.48\textwidth]{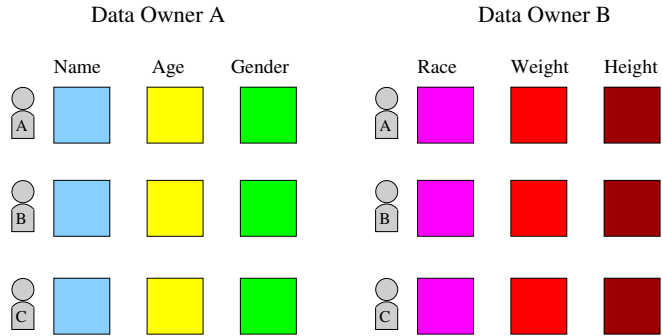}
\caption{An example of data partition in vertical federated learning.
In this example, 
data owners A and B consist of different feature space, 
while their data contains information about the same set of entities, 
A, B, and C.}
\label{fig:vfl-data}
\end{figure}

In this paper, 
we consider the privacy issues related to the vertical FL setting. 
Vertical federated learning is to divide the datasets vertically 
(by user feature dimension), then select partial data associated 
with the same set of users but user features are not exactly the same
for training. In other words, data in different columns represent the
same users. Therefore, vertical federated learning can increase the 
feature dimension of training data. Figure~\ref{fig:vfl-data} shows an 
example of feature separation between data owners in a vertical 
FL setting.

For example, there are two different institutions, one is a bank and
the other one is an e-commerce company. Their user groups 
contain an intersection set of users. However, because the bank has users’ 
income and expenditure behaviour and credit rating, while e-commerce 
keeps users’ browsing and purchasing history, their user features have
almost no intersection. 

Vertical federated learning is to aggregate these different features in an 
encrypted manner to enhance the model~\cite{yang2019}. At present, 
many machine learning models such as regression model~\cite{gascon2016}, 
tree structure model~\cite{vaidya2002,wu2020}, 
and neural network model~\cite{truex2019,zhu2021} have been applied to
this federated setting. Thus, protecting the privacy of the training data 
owned by each participating organisation should be a fundamental requirement in
vertical FL, as such training data might be highly sensitive~\cite{jiang2022}. 

Though, 
privacy-preserving techniques, 
such as differential privacy~\cite{Dwork2006} and  secure multiparty
computation (SMC)~\cite{cramer2015}, have been explored for horizontal
FL settings, however, only a handful of studies have been conducted to
explore the suitability of these techniques in vertical FL. Using SMC
and differential privacy can boost privacy protection in Federated
learning, but such protection often comes with a trade-off between cost 
and efficiency. 

When using SMC, each participating organisation can encrypt the 
parameters of its model before sending the model to a host organisation 
(aggregator). Therefore, additional computational resources are required
for encryption which will compromise the efficiency of training the 
model~\cite{bonawitz2017,truex2019}. With differential privacy, 
noise can be added to the model and data, at the cost of accuracy 
degradation~\cite{wei2020}. Thus, understanding and balancing the 
trade-off among privacy protection, operating efficiency, and model 
performance, both theoretically and empirically, are open and challenging 
problems in privacy-preserving vertical FL systems.

In this paper, we aim to explore how differential privacy can be 
applied to vertical FL systems. Differential privacy is the current 
state-of-the-art criterion to provide privacy protection with theoretical
guarantees. Under the differential privacy setting, we aim to explore the
effect of noise perturbation on model accuracy with different privacy budgets. 
To conduct experiments, we use several real-world datasets and evaluate
the model accuracy under different privacy budgets. Our experimental 
results show that with differential privacy noise we can trade the 
performance degradation of the vertical FL model for a certain level 
of privacy protection with calibrated noise perturbation.

This paper is organised as follows. In Section~\ref{sec:relwork} we 
discuss the current literature of vertical federated learning, and provide
preliminaries in Section~\ref{sec:background}. We describe the considered
vertical FL system in Section~\ref{sec:vfl-model}. We present and discuss
the results of our experiments in Section~\ref{sec:experiments}. Finally, 
we conclude and point out directions of future research in 
Section~\ref{sec:conclusion}. 

\section{Related works}
\label{sec:relwork}

Federated learning is actually a kind of encrypted distributed machine learning
technology, in which participants can build a model without disclosing the 
underlying data~\cite{mcmahan2017}. Through the parameter exchange under a secure 
communication mechanism, a common global model is established. Under such a mechanism, 
all parties involved can successfully use their data to build a machine learning 
model collaboratively~\cite{aledhari2020}.

The federated setting poses novel challenges to existing privacy-preserving 
algorithms. Beyond providing rigorous privacy guarantees, it is necessary to 
develop methods that are computationally cheap, communication efficient, 
and tolerant to dropped devices all without overly compromising 
accuracy~\cite{li2021, gong2020}.
Although there are a variety of privacy definitions in federated learning. 
However, the privacy in federated learning protocols can be considered under 
two scenarios, (1) the model updates generated at each iteration are private 
to all data owners other than the central server (trusted), while (2) the model 
updates are also private to the central server (untrusted or semi-trusted). 

The privacy techniques used in federated learning models can be typically 
categorised as secure multiparty computation and differential privacy.
Bonawitz et al.~\cite{bonawitz2017} used a secure multiparty computation 
based aggregation protocol in a horizontal FL model to protect individual
model updates from an untrusted central server. Sotthiwat et al.~\cite{sotthiwat2021} 
encrypt critical part of model (gradients) parameters to reduce communication cost, 
while maintaining MPC’s advantages on privacy-preserving without sacrificing 
accuracy of the learnt global model. Madi et al.~\cite{madi2021} a secure 
framework for verifiable FL relying on Homomorphic Encryption and Verifiable 
Computation. Recently, Kanagavelu et al.~\cite{kanagavelu2022} proposes a 
hierarchical model aggregation to reduce the communication cost incurred in MPC
enabled Federated Learning.

Due to the popularity, differential privacy is often use to enhance 
the privacy of each data owner in a federated learning settings~\cite{geyer2017,mohammadi2021,mcmahan2018,wei2020,kharitonov2019,andrew2021}. 
Geyer et al.~\cite{geyer2017} proposed a federated optimisation algorithm 
with differential privacy, which is applied to clients to ensure their global
differential privacy. Yu et al.~\cite{yu2021} efficient FL protocol which
protects the privacy of the IoT devices during the training process. 
McMahan et al.~\cite{mcmahan2017} proposed a protocol that applies 
differential privacy to federated learning and offer global differential 
privacy. Liu et al.~\cite{liu2021} proposed an asynchronous FL model which 
used local differential privacy to protect the client privacy while reducing
communication overhead during the training process. In order to avoid blindly 
adding unnecessary noise, Andrew et al.~\cite{andrew2021} designed a pruning
scheme based on adaptive gradient to reduce the penetration of noise to the 
gradient. Though, differential privacy provides efficient privacy solutions
compared to SMC techniques, such solutions bring uncertainty into the upload
parameters and may harm the training performance.

To achieve stronger privacy guarantees, differential privacy combined with
secure multiparty computation is used in  federated learning models.
Truex et al.~\cite{truex2019} proposed a federated learning model which 
utilises differential privacy with SMC  to reduce the growth of noise 
injection as the number of parties increases without sacrificing privacy
while maintaining a pre-defined rate of trust. 
Mugunthan et al.~\cite{mugunthan2019} a new mechanism that distributed 
differentially private noise utilization in a multi-party setting for
federated learning to reduce gradient leakage.

Though many of the techniques above proposed for horizontal FL, only few
techniques have been proposed for vertical FL scheme. 
Hardy et al.~\cite{hardy2017} proposed a vertical federated learning
model which uses a distributed logistic regression of Paillier additive 
homomorphic encryption scheme which can effectively protect privacy 
and also improve the accuracy of the classifier. In~\cite{yang2019a}, 
a quasi-Newton method was used for training logistic regression models 
in a vertical FL scheme to reduce the communication complexities.

Apart from these logistic regression models, few work have been proposed
for tree based models~\cite{cheng2021,liu2020a, hou2021, wu13}. 
Cheng et al.~\cite{cheng2021} proposed a lossless vertical federated 
learning scheme in which all parties combine user features to train
together to improve the accuracy of decision making. Liu et al.~\cite{liu2020a}
proposed a vertical FL scheme, called Federated Forest, which uses a tree based
prediction algorithm that largely reduce the communication overhead
while improving the prediction efficiency. 

Recently, Feng et al.~\cite{feng2020} a multiple data owners multi-class 
vertical FL framework that enables label sharing of each owner with other
participants in a privacy-preserving manner. In~\cite{hu2019,chen2020} 
asynchronous vertical FL frameworks are proposed where the local models are
updated by each party in an asynchronous manner and do not require feature 
sharing between parties.

As we have discussed above, vertical FL frameworks commonly uses
SMC technologies such as encryption to ensure secure and private learning. 
Further, to reduce computational complexities in the SMC techniques, few recent
works have utilised differential privacy (DP) into the training process to 
provide strict privacy guarantees for data of each data 
provider~\cite{cheng2021, wang2020}. However, DP based mechanisms tend to decrease 
utility in training. Thus, it is important to understand how to add the right 
amount of noise with out compromising the performance of the vertical FL model.

\section{Background}
\label{sec:background}

\subsection{Privacy threats in Vertical Federated Learning}
Recent studies have demonstrated that federated learning is vulnerable to 
multiple types of inference attacks, such as membership inference, property 
inference, and feature inference. However, membership inference is not 
meaningful in vertical FL as every participating data owner already knows 
the training sample IDs intrinsically during the sample alignment step. 
Thus, it is important to investigate other meaningful attacks that are 
applicable in a VFL setting. 

Luo et al.\cite{luo2021} studied the privacy leakage problem in the 
prediction stage of VFL, by presenting several feature inference attacks
based on model predictions. In the proposed attack model, the authors 
assumed that the adversary can control the trained vertical FL model and 
the model predictions. The model predictions can leak considerable information
about the features held by the data owners, which calls for designing 
private algorithms to protect the prediction outputs.

Jin et al.~\cite{jin2021} proposed a data leakage attack to efficiently
recover batch data from the shared aggregated gradients. Their experimental
results on vertical FL settings demonstrated their attack can perform 
large-batch data leakage effectively, thus VFL has a high risk of data 
leakage from the training gradients compared to its horizontal counterpart. 
The authors suggested that leveraging fake gradients in the training process
can overcome such privacy issues with a negative impact on the model 
performance.

Weng et al.~\cite{weng2020} proposed a reverse multiplication attack 
for logistic regression based VFL models~\cite{cheng2021,liu2020a, hou2021, wu13}. 
In this attack, the adversary reverse-engineers each multiplication term 
of the matrix product, so as to infer the target participant’s raw training 
data. As a countermeasure, the authors proposed to apply differential noise 
to protect the target participant’s private sensitive data set, but still will
be able to infer the information about the data set by training an 
equivalent model.

Sun et al.~\cite{sun2021} studied how to defend against input leakage
attacks in Vertical FL. The authors proposed a framework that stimulates 
a game between an attacker who actively reconstructs raw input from the
embedding layer and a defender who aims to prevent input leakage. The 
framework uses a noise regularisation module, which adds Gaussian noise 
to the training samples. Their experimental results show the suggested 
framework can effectively protect the privacy of input data while 
maintaining a reasonable model utility.

Recently, Fu et al.~\cite{fu2022} proposed three different label 
inference attack methods in the VFL setting: passive model completion, 
active model completion, and direct label inference attack. In model completion
attacks, the attacker needs extra auxiliary labelled data for fine-tuning 
its local model. Zou et al.~\cite{zou2022} proposed to use autoencoder and 
entropy regularization to hide the true labels as a countermeasure to
gradient inversion attacks and label inference attacks.

\subsection{Differential Privacy}

Differential privacy (DP)~\cite{dwork2014} is a privacy definition 
that guarantees the outcome of a calculation is insensitive to any 
single record in the data set. Differential privacy requires the output
of a data analysis mechanism to be approximately the same if any single
record is replaced with a new one. In order to achieve this privacy
guarantee, a DP algorithm must contain some form of randomness such 
that the probability of obtaining a particular outcome $o\in O$ from 
database $D$ is associated with any pair database-outcome $(D, o)$. 
Formally:
\begin{definition}[Neighbouring databases]
\label{def:neg-db}
Databases $\mathbf{D}\in\mathcal{D}$ and $\mathbf{D'}\in\mathcal{D}$ over 
a domain $\mathcal{D}$ are called neighbouring databases if they differ in 
exactly one record.
\end{definition}

\begin{definition}[Differential Privacy~\cite{Dwork2006}]
\label{def:dp}
A randomised algorithm $\mathcal{A}$ is $(\epsilon, \delta)$-differentially 
private if for all neighbouring databases 
$\mathbf{D}$ and $\mathbf{D'}$, and for 
all sets $\mathcal{O}$ outputs, we have
\begin{equation}
    Pr[\mathcal{A}(\mathbf{D})\in\mathcal{O}]\le
    exp(\epsilon)\cdot Pr[\mathcal{A}(\mathbf{D'})\in\mathcal{O}] + \delta,
\end{equation}
where $Pr[\cdot]$ denotes the probability of an event.
\end{definition}

We use the term \emph{pure differential privacy} when $\delta = 0$ and the 
term \emph{approximate differential privacy} when $\delta > 0$, 
in which case $\delta$ is typically a negligible value in the order of 
the inverse of the database size $|D|$.

Following this definition, 
adding an additional data point to a dataset must not substantially change 
the result of the algorithm $\mathcal{A}$. 
The formula is expressed using probabilities to account for the randomness 
in $\mathcal{A}$. If one guesses the results of the algorithm $\mathcal{A}$
to be in a set of possible values $\mathcal{O}$, then adding one data point
should not change the probability of being correct by more than $e^\epsilon$. 
If this is not the case, then adding the new data point might break the 
privacy promise because it is noticeable whether the data point was used 
by the query or not.

The amount of noise necessary to ensure differential privacy for 
$\mathcal{A}$  depends on the sensitivity of the algorithm $\mathcal{A}$. 

\begin{definition}[Sensitivity~\cite{Dwork2006}]
\label{def:sensitivity}
The sensitivity S(A) of algorithm A describes by how much the outputs 
can differ if the query is executed on two adjacent databases,
\begin{equation}
    S(A) = \underset{{\mathbf{D}, \mathbf{D'}}}{max} ||\mathcal{A}(\mathbf{D}) - \mathcal{A}(\mathbf{D'})||_2,
\end{equation}
where $||\cdot||_2$ is the $L_2$-norm. 
\end{definition}

\section{Vertical Federated Learning Model}
\label{sec:vfl-model}

We now describe the vertical federated learning (VFL) model in detail. 
As we explained before, 
the basic idea of FL is to collaboratively train a machine learning 
model by a group of data owners, where only model updates are shared 
during the training process and local data never leaves its owner. 

In the VFL setting, each data owner holds different features of the 
same set of samples. Thus, the main goal of VFL is to aggregate these 
different features in an encrypted state to enhance the performance of 
the model. Besides the local training organisations, we assume a host 
organisation that holds the labels of these training samples. During 
the inference phase, the host organisation is responsible for privately
collecting and aggregating intermediate results of all data owners, 
computing the confidence scores, and sending them back to the local 
organisations for updating their models. 

\begin{figure}[!t]
\centering
\includegraphics[width=0.48\textwidth]{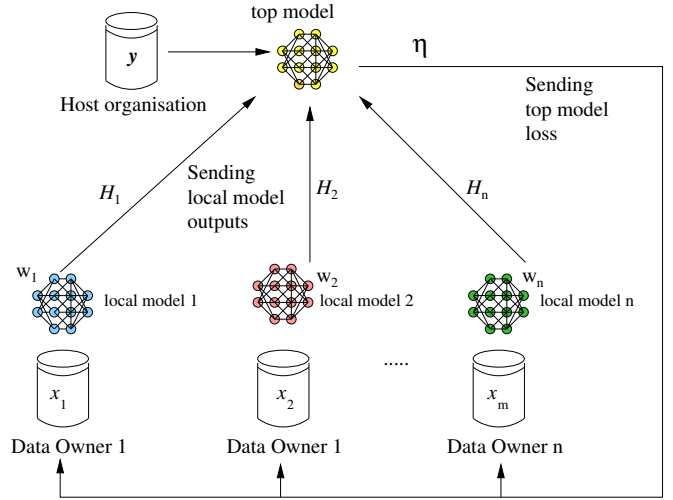}
\caption{An illustration of the vertical federated learning protocol.}
\label{fig:vfl-train}
\end{figure}

We define the VFL training process formally for $n$ data owners 
as follows and this training process is illustrated in 
Fig.~\ref{fig:vfl-train}. Let us assume a data set 
$\mathcal{D} = \{x, y\}$ with $\{x\}$ and $\{y\}$ being a set of 
feature vectors and the corresponding label data, respectively. 
We assume the set of feature vectors $\{x\}$ can be decomposed into 
$n$ blocks with $\{x\} = \{x_i\}_{i=1}^n$ with each block $\{x_i\}$ 
held by data owner $i$. We assume that the set of labels ${y}$ is 
owned by the host organisation. This is a reasonable assumption because
the labelled data contains valuable information and is not easy to 
come by in practice~\cite{yang2019}. 

Following~\cite{liu2021a}, 
each data owner $i$ trains a sub-model $\mathcal{M}_i$ which outputs 
local latent embedding $E_i$ represented by,
\begin{equation}
    E_i=\mathcal{M}_i(w_i, \{x_i\}),
\end{equation}
where $w_i$ is the training parameters of data owner $i$.
Here, 
$\mathcal{M}_i$ can take different forms of models such as linear 
and logistic regression, support vector machines, neural networks. 

Each party sends the latent embedding $E_i$ to the host organisation, 
which concatenates these embeddings into the final embedding $\mathcal{E}$, 
$\mathcal{E} = concat(E_i)_{i=1}^n$. 
The host organisation feeds the final embedding into their training 
model (named \emph{top model}) $\mathbf{\mathcal{M}}$, which outputs the 
predictions ${\Theta}(\mathcal{E})$ where ${\Theta}(\cdot)$ is a non-linear
operation such as sigmoid or softmax function. Thus, we can define a 
classification loss function $l(\cdot)$ for this training process as, 
\begin{equation}
    l(w_i, w_2, \cdots, w_n; \mathcal{D}) = l(\Theta(\mathcal{E}), y).
\end{equation}

Without loss of generality, 
the collaborative training problem can be formulated as,
\begin{equation}
    \underset{W}min(W, \mathcal{D} ) \triangleq l(\Theta(\mathcal{E}), y)\ + 
    \lambda(\sum_{i=1}^n \gamma(w_i)),
\end{equation}
where $W=[w_i, w_2, \cdots, w_n]$, $\lambda$ is the hyperparameter, 
and $\gamma(\cdot)$ is the regularizer. 
Finally, 
the host organisation calculates the gradient update 
$\eta=\frac{\partial l}{\partial \mathcal{E}}$ and sends it to the data owners. 
Thus, the gradient update of each data owner $i$ in the VFL framework 
can be computed as,
\begin{equation}
    \nabla_n\ l(w_1, w_2, \cdots, w_n; \mathcal{D}) = 
    \frac{\partial l}{\partial \mathcal{E}}\frac{\partial E_i}{\partial w_i}.
\end{equation}

\subsection{Threat model} 
We next describe the threat model we consider in this VFL setting.
We assume all data owners are honest while the host organisation is 
honest-but-curious. Thus, the host organisation follows the steps of the 
VFL protocol but is trying to infer as much as possible about the data
of data owners. We assume that the host organisation does not control 
any data owner's training or the global communication protocols. 
In addition, we do not assume the attacker has auxiliary information about
the data of each individual data owner, which can be difficult or 
impossible to obtain in real-world applications. 

\subsection{The application of differentially private noise}

As we explained in Section~\ref{sec:background}, the privacy of VFL
is challenged by various new attacks. Thus, we aim to investigate 
how differential privacy can be used in the training process as a 
countermeasure for such attacks. 
 
We assume $\mathcal{T}$ iterations are used in the VFL training process.
Following the threat model described above, in each training iteration, 
we need to hide the information about the data of a specific user is 
either used or not used at all for training from the host organisation. 
To account for this, following Definition~\ref{def:neg-db}, we consider
two training samples $\mathbf{D}$ and $\mathbf{D'}$ to be adjacent 
ones such that they only differ in the data of one single user. 
That is, the data samples are the same except that one of the samples
contains data from a user that is not present in the other data sample. 
The intuition is that it will be difficult for the host organisation to
learn whether a user participated in training the model. Thus, 
we assume the VFL model should not differ much by adding a new user. 
 
Following Abadi et al.~\cite{abadi2016}, 
we introduce the following theorem to design a machine learning 
algorithm, which can be proven to be $(\epsilon, \delta)$-differentially
private based on a noisy stochastic gradient decent (SGD) mechanism. 

\begin{theorem}
\label{thm:dp}
A machine learning algorithm $\mathcal{A}$ based on stochastic 
gradient descent computes a gradient estimate in each of 
$\mathcal{T}$ training iterations. The data used to compute the 
estimate is sampled using a probability $p$. The sensitivity of the 
estimate is bounded by a constant $d$ and we can add Gaussian noise
sampled from $N(0, \sigma^2 d^2)$ to the gradient estimate in each 
iteration. To compute the weights (training parameters) $w_{j+1}$ of
the next iteration $j+1$, the estimate is subtracted from the current
weights $w_{j}$ of a given iteration $j$. Thus, if there exists 
constants $c_1$ and $c_2$, then the algorithm $\mathcal{A}$ is 
$(\epsilon, \delta)$-differentially private for any
$\epsilon < c_1 \sigma^2 \mathcal{T}$ with $\delta>0$ and the 
standard deviation of noise is characterized by:
\begin{equation*}
    \sigma \le c_2 \frac{p\sqrt{\mathcal{T}log(1/\delta)}}{\epsilon}.
\end{equation*}
\end{theorem}

Following~\cite{mcmahan2018}, 
this can be adapted to a VFL setting. 
In each iteration of FL, 
the data of each data owner $i$ is sampled with a probability of $p$. 
This ensures that the number of sampled records can differ across 
iterations. To bound the sensitivity of the gradient estimate, 
we can bound the size $s$ that an individual local model update $w_i$ 
can have. Thus, this can be implemented by checking the $L_2$ norm of 
$w_i$, i.e., 
\begin{equation}
     \tilde{w_i} =
  \begin{cases}
    w_i       & \quad \text{if } ||w_i||_2 \le s \\
    w_i \times \frac{s}{||w_i||_2}  & \quad \text{otherwise}
  \end{cases}
\end{equation}
For a neural network model with $K$ layers, 
the overall limit $s$ can be computed as,
\begin{equation}
    s = \sqrt{\sum_{k=1}^K s_k},
\end{equation}
where $s_k$ is the limit of the $L_2$-norm in the $k$-th layer. 

If the set of sampled data is denoted by $\mathbf{C}$, 
then we can estimate the gradient as,
\begin{equation}
    \mathbb{E}_g(\mathbf{C}) = \frac{\sum_{j\in \mathbf{C}}m_j\times w_j}{\sum_{j\in \mathbf{C}}m_j}, 
\end{equation}
where the number of data points $m_j$ reflects the importance of 
the selected data sample in iteration $j$. Assuming 
$N=\sum_{j\in \mathbf{C}}m_j$ is the number of records in the current sample, 
we can estimate the $L_2$-norm of estimated gradients as, 
\begin{equation}  
\begin{split}
||\mathbb{E}_g(\mathbf{C})||_2 & = 
||\frac{\sum_{j\in \mathbf{C}}m_j\times w_j}{\sum_{j\in \mathbf{C}}m_j}||_2 \\
 & = ||\sum_{j\in \mathbf{C}}\frac{m_j}{N}w_j||_2 \\
 & \le \sum_{j\in \mathbf{C}}||\frac{m_j}{N}w_j||_2\\
 & = \sum_{j\in \mathbf{C}}\frac{m_j}{N}||w_j||_2\\
 & \le \frac{\sum_{j\in \mathbf{C}}m_j}{N}s\\
 & = s.
\end{split}
\end{equation}
This can be expanded into the calculation of sensitivity bounds 
of gradient estimates as, 
\begin{equation}
    \begin{split}
        S(\mathbb{E}_g) & = \underset{\mathbf{C}, \mathbf{D}}{max} ||\mathbb{E}_g(\mathbf{C}) - \mathbb{E}_g(\mathbf{C}\cup\mathbf{D}))||_2\\
        & \le \underset{\mathbf{C}, \mathbf{D}}{max}||\mathbb{E}_g(\mathbf{C})|| + ||-\mathbb{E}_g(\mathbf{C}\cup\mathbf{D}))||_2\\
        & = \underset{\mathbf{C}, \mathbf{D}}{max}||\mathbb{E}_g(\mathbf{C})|| + ||\mathbb{E}_g(\mathbf{C}\cup\mathbf{D}))||_2\\
        & = \underset{\mathbf{C}, \mathbf{D}}{max} 2s\\
        & = 2s.
    \end{split}
\end{equation}

This shows that the sensitivity of the gradient estimate is bounded 
which allows us to apply Theorem~\ref{thm:dp} in the VFL setting. 
Thus, one can apply a sufficient amount of Gaussian noise to model 
parameters in each iteration. However, it is challenging to choose 
the privacy budget ($\epsilon$) that provides an appropriate level 
of differential privacy. This is a common problem with differential 
privacy in general and applies to any federated learning system. 
Next, we empirically investigate the accuracy of the VFL training
under various privacy budgets.

\section{Experimental evaluation}
\label{sec:experiments}
In this section, 
we present our experimental results to demonstrate the usefulness 
of applying differential privacy in a vertical federated 
learning (FL) setting. We first describe the datasets and parameters
we used for the experiments. Then, we discuss our experimental results. 

\subsection{Datasets}

We used four different datasets in our experiments. 
We summarize their basic facts in Table~\ref{tab:datasets}.
In each of these datasets, we used 20\% records for testing and the 
remaining for training.

\begin{table*}[t!]
\centering
  \caption{Overview of the data sets used in the experiments. 
  \label{tab:datasets}}
\begin{small}
\begin{tabular}{lcccc}
\hline\noalign{\smallskip}
Dataset & Domain &
\begin{tabular}[c]{@{}l@{}}Number of \\ records\end{tabular} &
\begin{tabular}[c]{@{}l@{}}Number of\\ Attributes\end{tabular} &
Classification \\
\hline\noalign{\smallskip}
Adult     & Census  &  32,561 & 9     & Binary classification  \\
\hline\noalign{\smallskip}
Sport     & Sport   &  9,120  & 5,625 & Binary classification  \\
\hline\noalign{\smallskip}
Energy    & Energy  &  768    & 10    & Multi-output Regression \\
\hline\noalign{\smallskip}
Boston-Housing & Product &  506 & 12    & Regression \\
\hline\noalign{\smallskip}
California-Housing & Product &  20,640 & 8    & Regression \\
\hline\noalign{\smallskip}
\end{tabular}
\end{small}
\end{table*}

The adult dataset contains records of individuals from 1994 US Census, 
and is used to predict if an individual’s annual income exceeds 
50,000, which can be viewed as a binary classification problem. 
We use daily and sports activities dataset~\cite{altun2010} that 
comprises motion sensor data of 19 daily and sports activities, 
each performed by 8 subjects (4 female, 4 male, between the ages 
20 and 30) in 5 minutes. This dataset contains 9,120 samples and 
5,625 attributes. This dataset is also used for formulating a 
classification problem.

We also used an energy dataset~\cite{tsanas2012} 
which contains eight attributes (or features, denoted by X1…X8) 
and two responses (or outcomes, denoted by y1 and y2) of energy 
usage of 768 different building shapes.  Here, 
the eight features used are relative compactness, surface area, etc.
The two responses are heating load and cooling load. 
The aim is to use the eight features to predict each of the two
real-valued responses, thus making this dataset be used for
formulating a regression problem. 


The last two dataset we used in our experiments are the 
Boston-Housing dataset~\cite{harrison1978} and the 
California-Housing dataset~\cite{pace1997}. We used both of these
dataset to predict the median house values which can be viewed as a 
univariate regression problem.
The Boston-Housing dataset contains 506 records each with the features of 
capita crime rate by town, proportion of residential land zoned for lots over
25,000 square feet, proportion of non-retail business acres per town, 
Charles River dummy variable (= 1 if tract bounds river; 0 otherwise),
Nitric Oxides concentration (parts per 10 million),
average number of rooms per dwelling,
proportion of owner-occupied units built prior to 1940,
weighted distances to five Boston employment centres,
index of accessibility to radial highways,
full-value property-tax rate per USD 10,000,
pupil-teacher ratio by town, and
lower status of the population.

The California-Housing dataset was derived from the 1990 U.S. census, 
using one row per census block group. A block group is the smallest 
geographical unit for which the U.S. Census Bureau publishes sample 
data (a block group typically has a population of 600 to 3,000 people).
This dataset contains features of median income in block group,
median house age in block group, average number of rooms per household,
average number of bedrooms per household, block group population,
average number of household members, block group latitude, and 
block group longitude.

\subsection{Parameter settings}

In our experiments, we assumed 3, 4, and 5 data owners with one 
host organisation holding the labels. For each of the data set, 
we split the number of attributes equally among the data owners. 

We adopt a classic multilayer perceptron (MLP) network consisting
of three hidden layers with 48, 96 and 196 neurons, respectively.
The second layer is chosen as the cut/splitting layer, where the
size of the forward output of each bottom model will be 32 and
the input size of the top model will be 96. We adopt an SGD 
optimizer and the learning rate is set to 0.0002. We set the number
of training epochs (iterations) to 100 and used batch processing
with a batch size of 100. 

\begin{figure}[!t]
\centering
\includegraphics[width=0.24\textwidth]{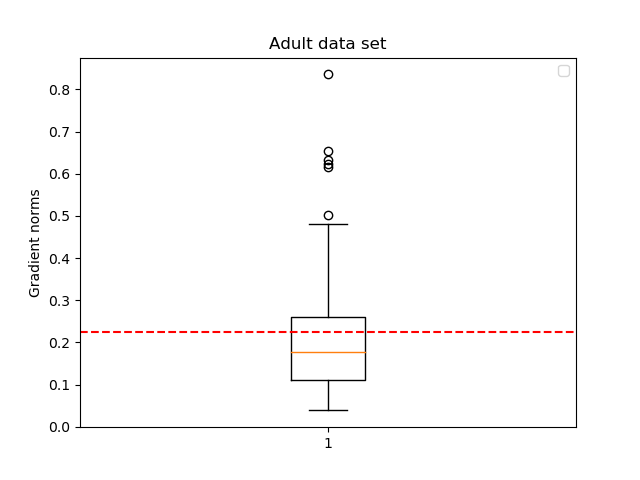}
\includegraphics[width=0.24\textwidth]{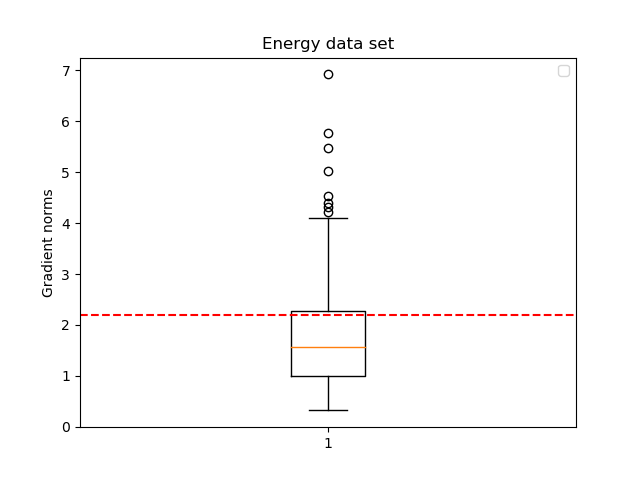}
\caption{Clipping threshold selection for Adult and Energy datasets.}
\label{fig:clip_threshold}
\end{figure}

Following Definition~\ref{def:dp}, 
we applied Gaussian noise with $\delta$ to 0.001 to local model 
parameters (model weights) of each data owner before sending these 
parameters to the host organisation. 
We consider sequential composition over training cycles where 
differential privacy noise is added to model weights. 
To prevent gradients from explosion due to noise addition, 
we used gradient clipping with a clipping threshold. 
We set the clipping threshold to the 80th percentile of the gradient 
norms of the first 50 iterations. Figure~\ref{fig:clip_threshold} shows
the clipping threshold for Adult and Energy datasets. 
We set the privacy budget ($\epsilon$) value to 1, 1.5, 2, 5, 10, 50, 
and 100. 

For performance evaluation, we used test accuracy, test loss, and 
mean square error as metrics. As baselines, we selected two approaches. 
First, we adopted a centralised learning setting without adding any 
differential noise. This baseline provides the conventional learning 
setting where a single data owner holds the training data and their
labels. As the second baseline approach, we used the federated 
learning setting without adding any differential noise in the model 
parameters. This baseline shows the best utility that can be achieved
for each dataset in our VFL setting. 

We implemented all approaches in Python 3.7 and we used TensorFlow to 
implement our deep learning model~\cite{abad16}. 
All experiments were performed on a 64-bit Intel Core i9 chip, 
with eight cores running 16 threads at speeds of up to 2.4GHz, 
along with 64 GBytes of memory, and running on Windows 10. 
To facilitate repeatability, the data sets and the programs will be 
made available to the readers. 

\subsection{Results and Discussion}

\begin{figure*}[!t]
\centering
\includegraphics[width=0.3\textwidth]{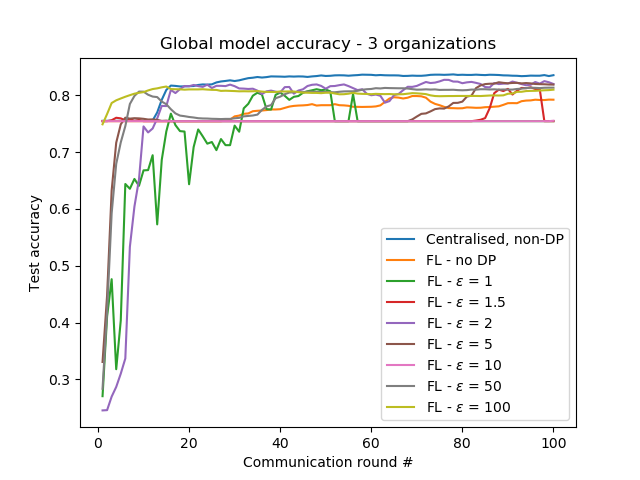}
\includegraphics[width=0.3\textwidth]{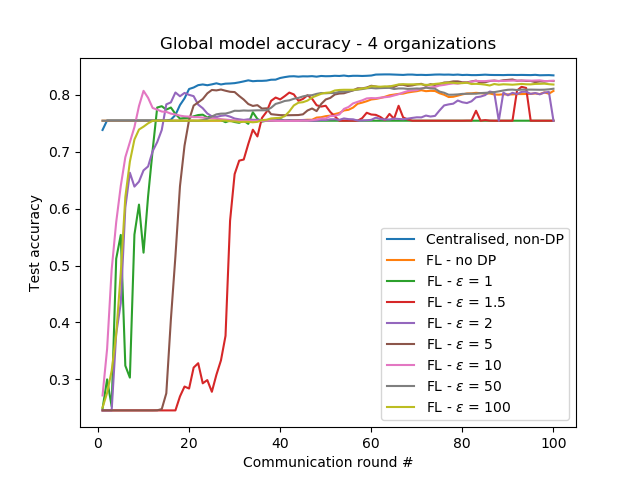}
\includegraphics[width=0.3\textwidth]{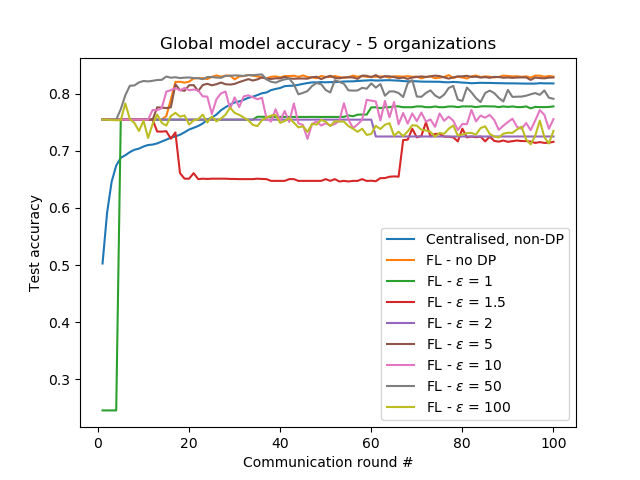}
\includegraphics[width=0.3\textwidth]{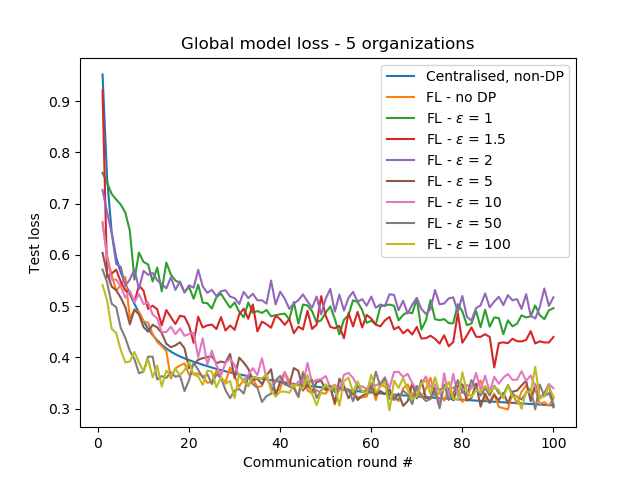}
\includegraphics[width=0.3\textwidth]{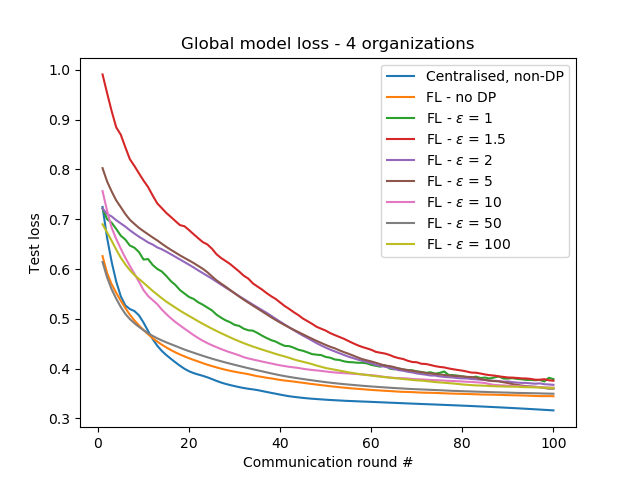}
\includegraphics[width=0.3\textwidth]{figures/loss_ADULT_5.png}
\caption{Test accuracy (top row) and loss (bottom row) for Adult 
         dataset with different numbers of data owners.}
\label{fig:adult-results}
\end{figure*}

\begin{figure*}[!t]
\centering
\includegraphics[width=0.3\textwidth]{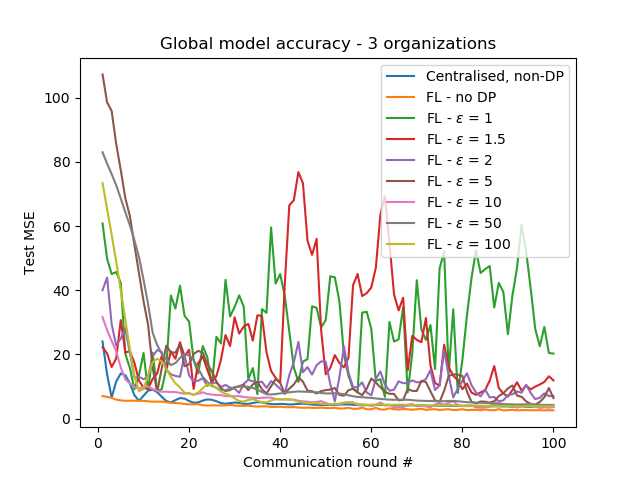}
\includegraphics[width=0.3\textwidth]{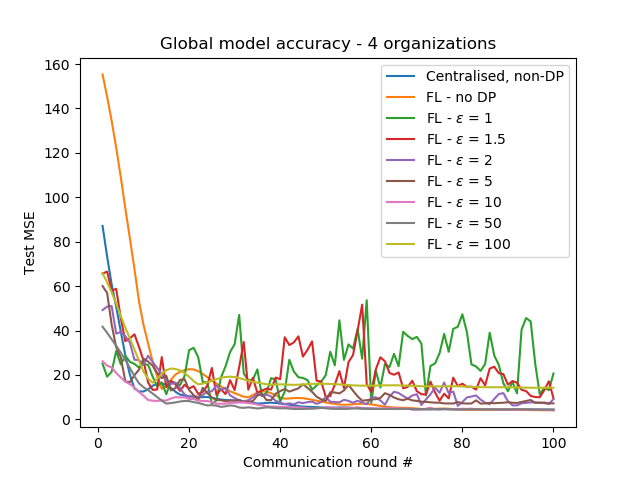}
\includegraphics[width=0.3\textwidth]{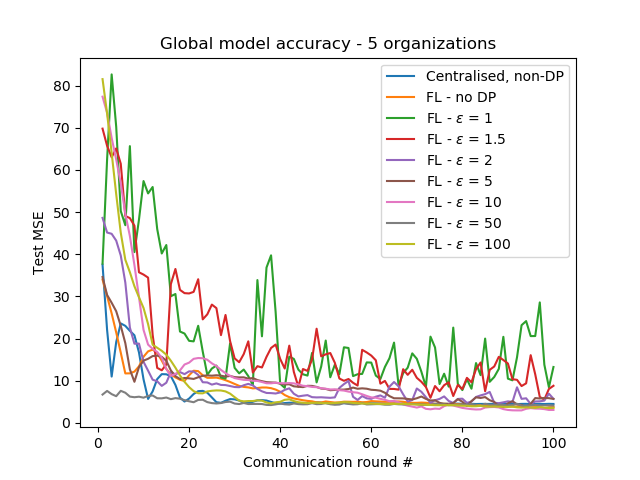}
\includegraphics[width=0.3\textwidth]{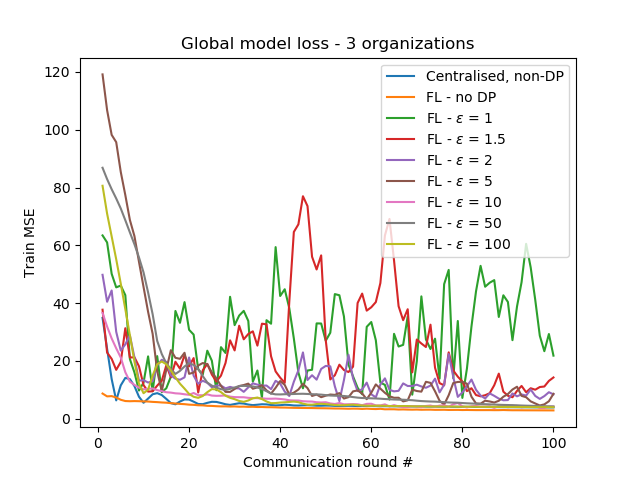}
\includegraphics[width=0.3\textwidth]{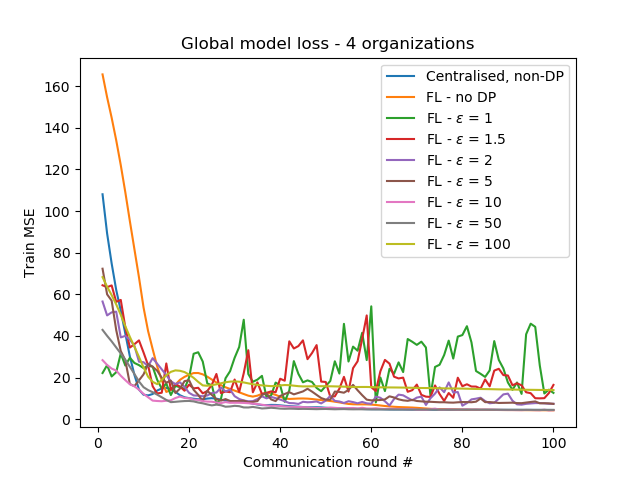}
\includegraphics[width=0.3\textwidth]{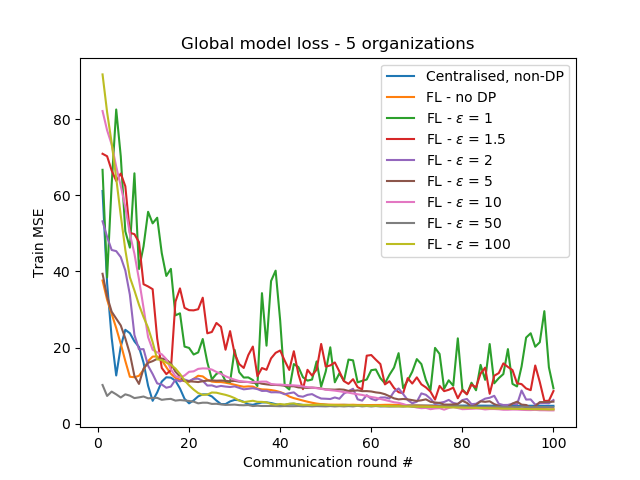}
\caption{Test accuracy (top row) and loss (bottom row) for Energy 
         dataset with different numbers of data owners.}
\label{fig:energy-results}
\end{figure*}

\begin{figure*}[!t]
\centering
\includegraphics[width=0.3\textwidth]{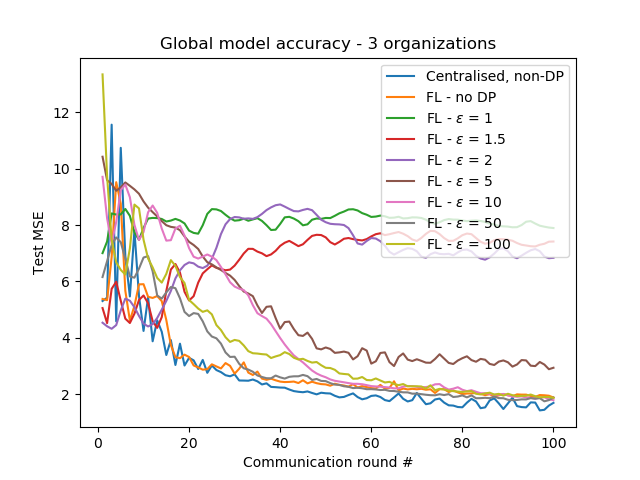}
\includegraphics[width=0.3\textwidth]{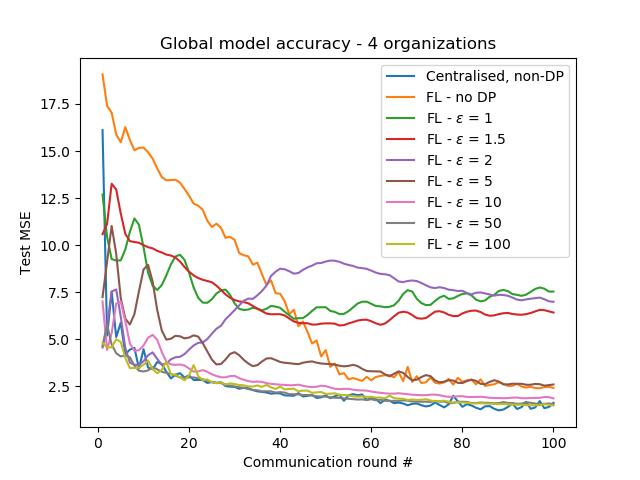}
\includegraphics[width=0.3\textwidth]{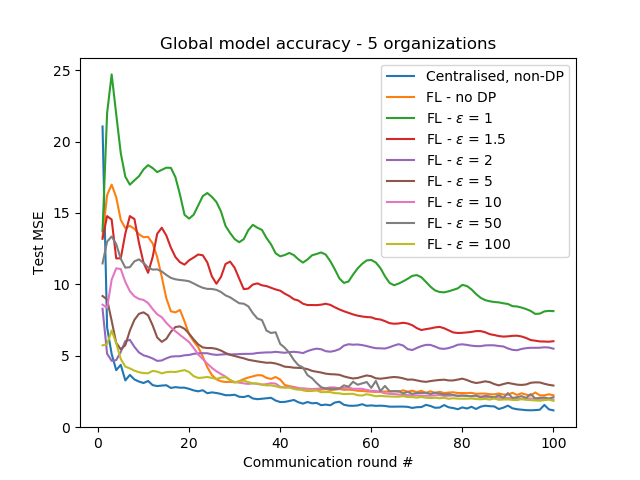}
\includegraphics[width=0.3\textwidth]{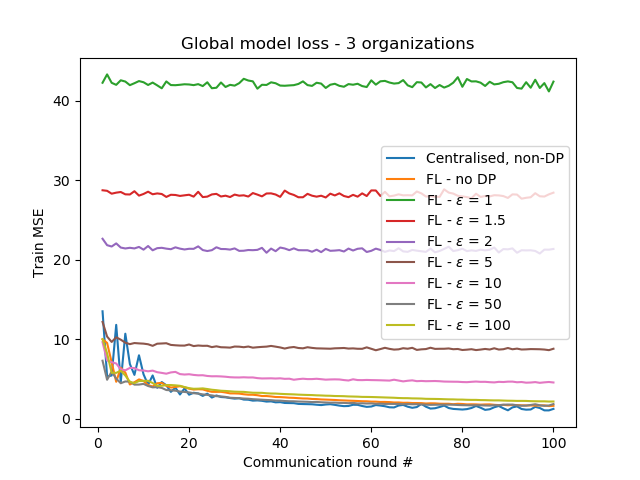}
\includegraphics[width=0.3\textwidth]{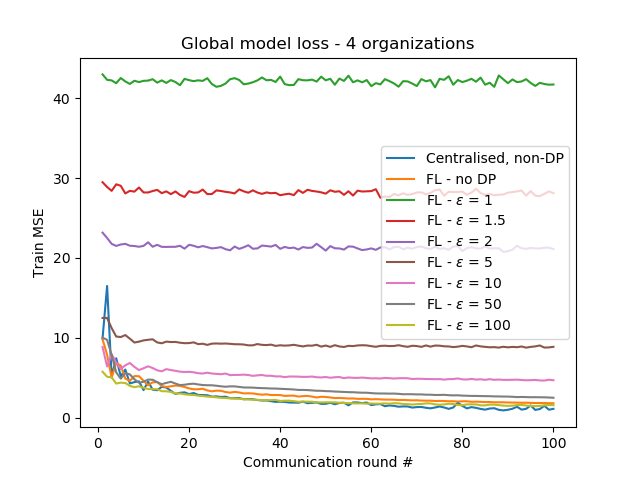}
\includegraphics[width=0.3\textwidth]{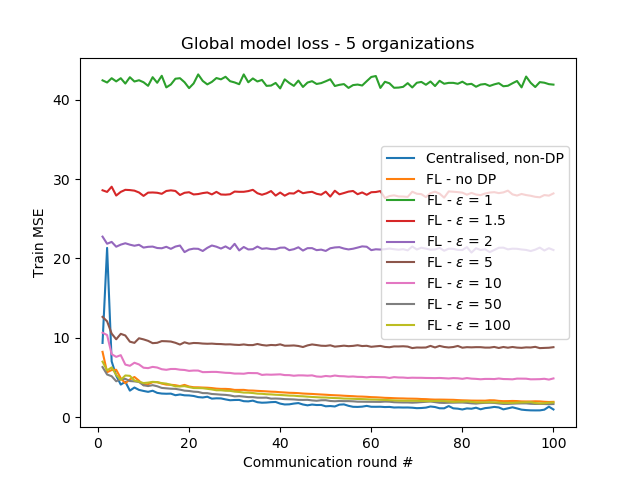}
\caption{Test accuracy and loss for Sport dataset with different numbers of data owners.}
\label{fig:sport-results}
\end{figure*}

\begin{figure*}[!t]
\centering
\includegraphics[width=0.3\textwidth]{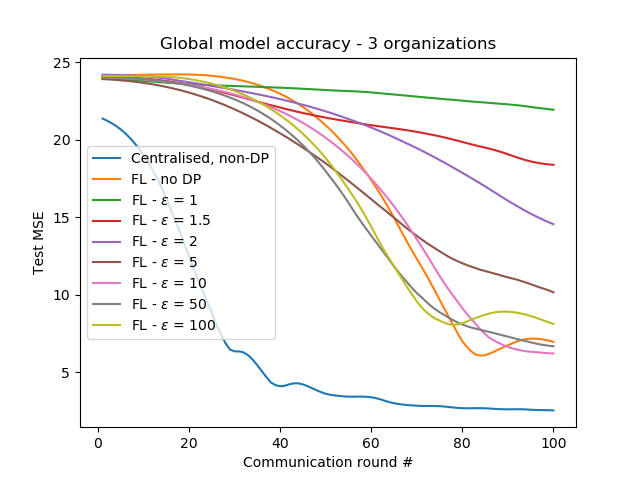}
\includegraphics[width=0.3\textwidth]{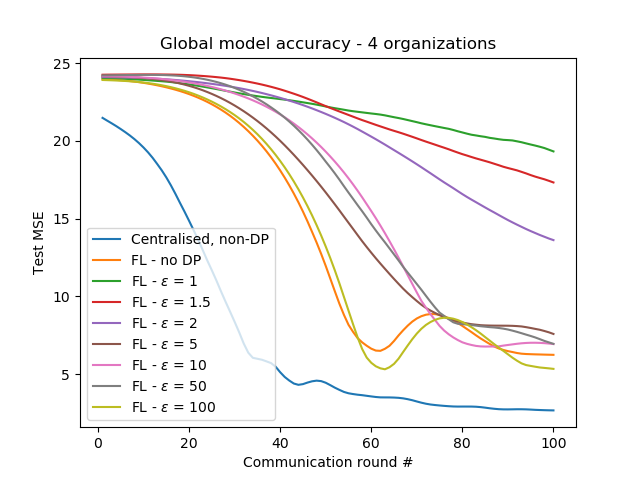}
\includegraphics[width=0.3\textwidth]{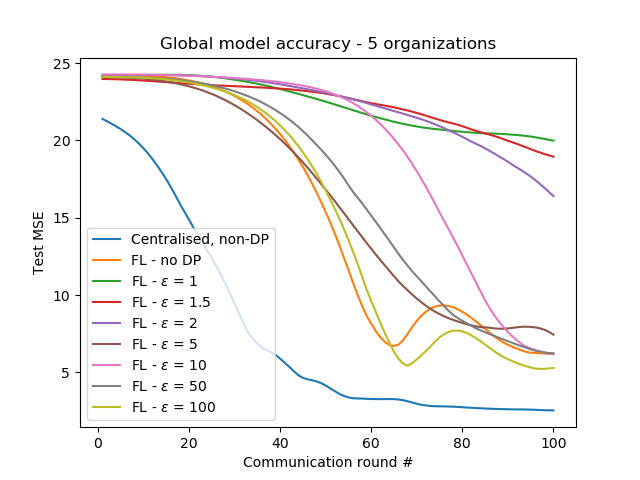}
\includegraphics[width=0.3\textwidth]{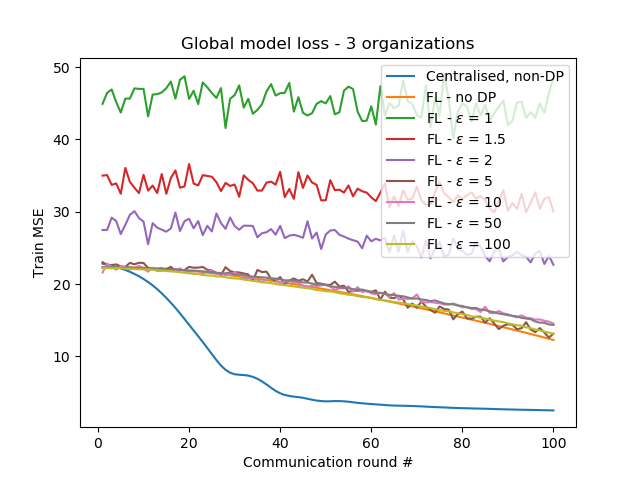}
\includegraphics[width=0.3\textwidth]{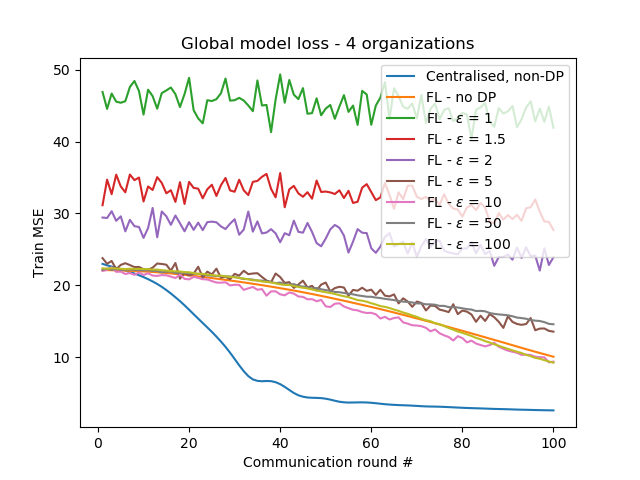}
\includegraphics[width=0.3\textwidth]{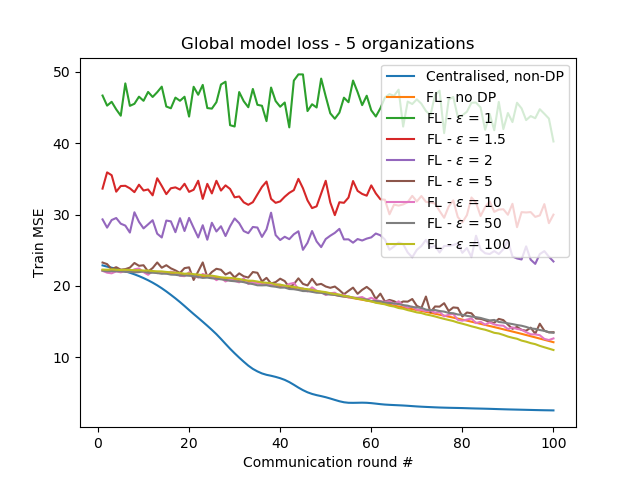}
\caption{Test accuracy (top row) and loss (bottom row) for Boston-Housing dataset with 
different data owners.}
\label{fig:boston-results}
\end{figure*}

\begin{figure*}[!t]
\centering
\includegraphics[width=0.3\textwidth]{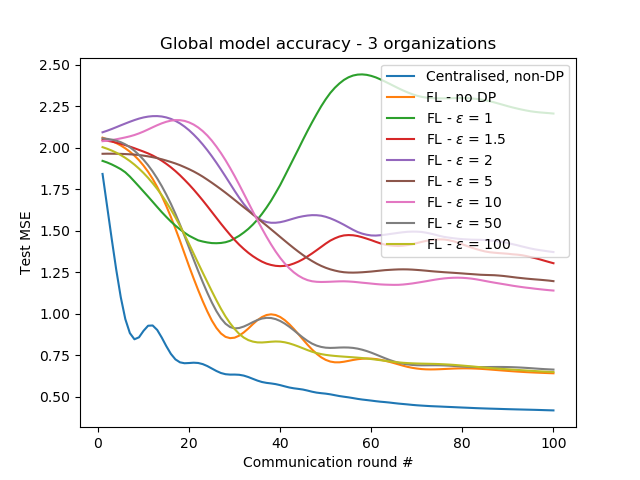}
\includegraphics[width=0.3\textwidth]{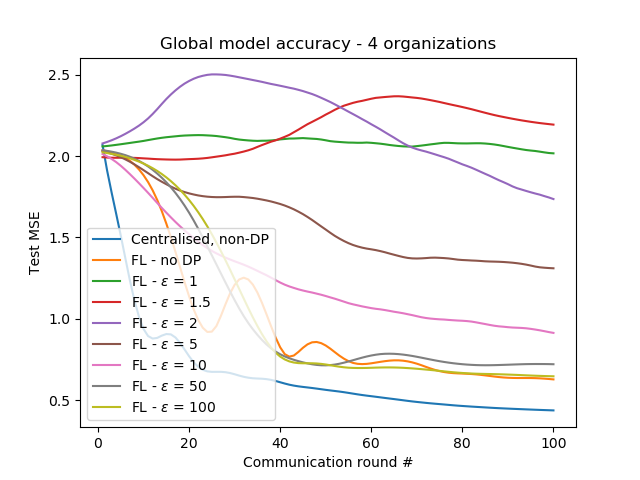}
\includegraphics[width=0.3\textwidth]{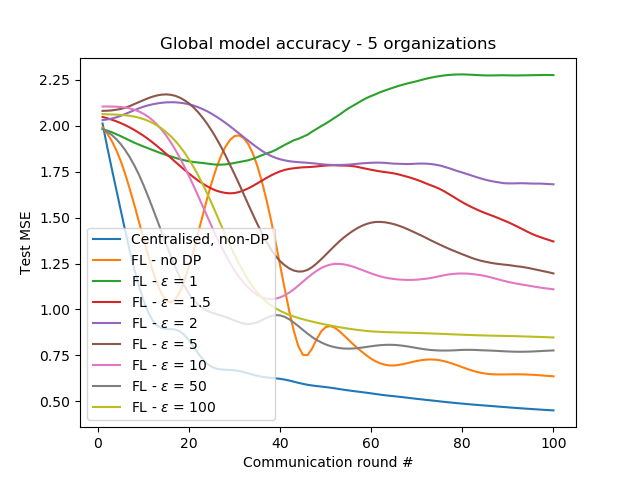}
\includegraphics[width=0.3\textwidth]{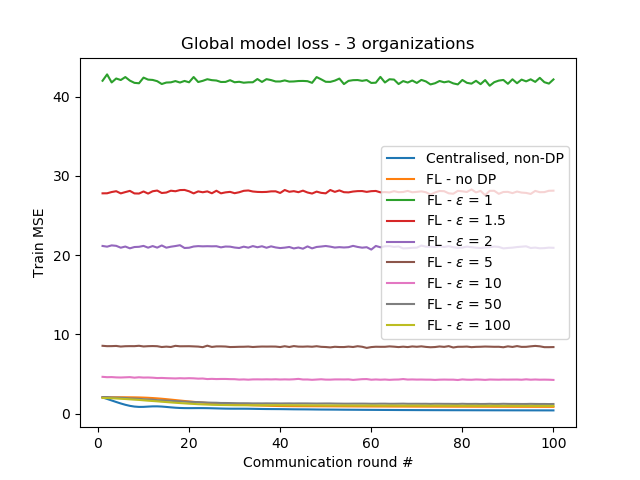}
\includegraphics[width=0.3\textwidth]{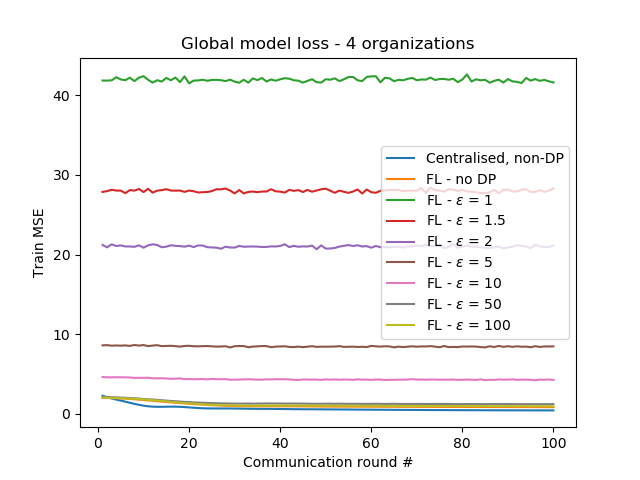}
\includegraphics[width=0.3\textwidth]{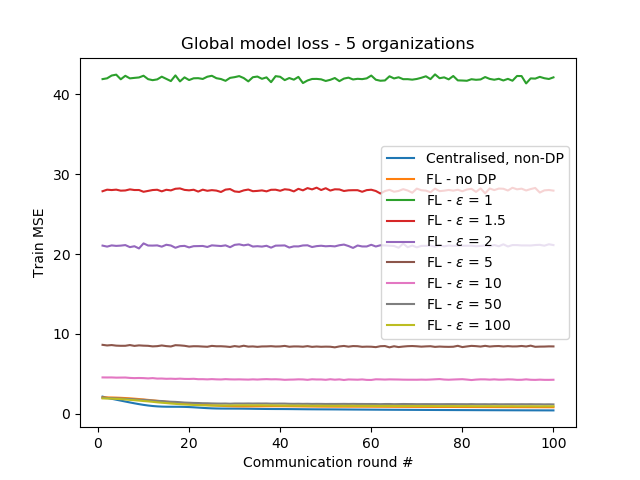}
\caption{Test accuracy (top row) and loss (bottom row) for California-Housing dataset with 
different data owners.}
\label{fig:california-results}
\end{figure*}

Figures~\ref{fig:adult-results} to~\ref{fig:california-results} show the
accuracy and loss results for different datasets. As can be seen from
those figures, the model accuracy is affected by the added 
differential privacy noise. When the models are perturbed with large
amount of noise (e.g. $\epsilon < 10 $), the overall accuracy of the
model is lower compared to the VFL setting without DP noise. This is 
due to the fact that the perturbed forward output will make the loss
function value biased. Therefore, the fundamental relationship between
the convergence bound and the privacy level needs to be characterised 
to achieve configurable trade-off requirements.

As expected, 
when the local model parameters are perturbed with less amount of 
noise, the training of the VFL setting tends to achieve similar 
classification accuracy as the centralised learning setting. 
Also, we noted that the results in terms of the loss of trained 
models do not change much after 20 epoch cycles for some datasets.

\section{Recommendations and Future Directions}

Federated learning is an active and ongoing area of research. Due to 
the growing interest in federated applications, some
recent work has begun to address some of the challenges that are unique in the
vertical federated settings. 
Several critical open questions are yet to be explored. In this section, we outline a few 
promising research directions surrounding the previously discussed privacy
challenges and introduce additional challenges regarding dataset 
characteristics.

\subsection{Privacy attacks and possible defences}

As we discussed in Section~\ref{sec:background}, VFL systems 
are vulnerable to different privacy attacks\cite{luo2021,jin2021,weng2020,fu2022}. 
Among those attacks, label inference attacks are powerful against VFL 
on real-world, large-scale datasets. Protecting the privacy of the 
labels owned by the host organisation should be a fundamental guarantee 
provided by VFL, as the labels might be highly sensitive. 
For example, in a medical application, 
the labels might indicate whether a person has a
certain kind of disease. 
However, as shown by Fu et al.~\cite{fu2022}
the bottom model structure and the gradient update mechanism of VFL
can be exploited by a malicious data owner to infer the privately owned
labels.

Defences, such as gradient compression and noisy gradients, 
are possible 
countermeasures that can effectively mitigate the threat of a direct label 
inference attack. However, such mechanisms are ineffective against
passive and active label inference attacks because an adversary can 
fine-tune the bottom model with an additional classification layer for 
label inference using a small amount of auxiliary labelled data. 
Thus, this calls for new defence strategies designed for VFL systems. 


\subsection{Vertical Federated Unlearning}

Recent legislation and regulations in the EU and US have established 
a new data privacy right called "the right to erasure" 
(e.g., in the EU's General Data Protection 
Regulation~\footnote{https://gdpr-info.eu/)}) or "the right to delete" 
(e.g., in the California Consumer Privacy 
Act~\footnote{https://oag.ca.gov/privacy/ccpa)}). This privacy right of
"data erasure" states that under many circumstances, a user can request 
an organization to erase his/her data entry from a database, even 
after data collection and analysis. Such a strong level of privacy 
protection can make a user utterly invisible in a data life cycle, 
thus safeguarding fundamental human rights such as freedom of association 
and freedom from discrimination. In the context of FL, 
"data erasure" means that the learning should be partially undone by 
forgetting the training data of a given local client. 
Such a reverse learning operation is recently referred to as 
"federated unlearning"~\cite{halimi2022federated}. However, existing work on
federated unlearning treats HFL only. Its extension to VFL is unclear and
non-trivial. More specifically, erasing one data sample from HFL is one
thing, but forgetting some attributes of multiple samples or even an 
entire data silo from VFL is another story.

\subsection{Ethical Concerns}
The paradigm of VFL opens up new avenues of data collection from a 
variety of sources for a host organization. 
However, the bright future of big data and AI analytics using VFL 
also raises ethical concerns. Take insurance pricing as an example; 
although direct discrimination is prohibited by legislation and 
regulations, indirect discrimination is a grey area because the pricing 
models and algorithms are usually opaque and not accessible.
More specifically, when applying VFL to pricing insurance contracts, 
would it be ethical and fair to use the consumer data of grocery/online shopping, 
fitness levels from gym surveys or wearable trackers, 
income/taxation/employment information, membership status in 
various groups, and household energy usage patterns from smart 
meters? If yes, would it be acceptable to further dip into the consumer
data of prescription medications, social media posts/images/videos, 
vehicle GPS trajectories, Youtube watching lists, or web browsing history? 
Where is the boundary of data usage in VFL for insurance underwriting 
and other applications? In practice, we believe multi-disciplinary 
research is needed to investigate VFL's discrimination and fairness 
issues for various business use cases
to reduce the risk of harm to consumers in terms of exclusion and prejudice. 

\section{Conclusion}
\label{sec:conclusion}
Recent advancements in federated learning (FL) allow multiple data
owners/organisations to collaboratively train a machine learning 
model without sharing their raw data. In this paper, we explore how
differential privacy can be used in a vertical federated learning 
setting. It is a widely adopted method to add DP noise to the output
in the process of gradient iteration, so as to achieve the goal of 
privacy protection. We can further add noise to the data to enhance
privacy protection. However, DP comes at a cost of sacrificing the
performance of the model. Hence, a reasonable trade-off point needs 
to be found to determine the appropriate amount of additive DP 
noise, while maintaining a useful model.

\bibliographystyle{IEEEtran}
\bibliography{main}

\begin{thebibliography}{10}
\providecommand{\url}[1]{#1}
\csname url@samestyle\endcsname
\providecommand{\newblock}{\relax}
\providecommand{\bibinfo}[2]{#2}
\providecommand{\BIBentrySTDinterwordspacing}{\spaceskip=0pt\relax}
\providecommand{\BIBentryALTinterwordstretchfactor}{4}
\providecommand{\BIBentryALTinterwordspacing}{\spaceskip=\fontdimen2\font plus
\BIBentryALTinterwordstretchfactor\fontdimen3\font minus
  \fontdimen4\font\relax}
\providecommand{\BIBforeignlanguage}[2]{{%
\expandafter\ifx\csname l@#1\endcsname\relax
\typeout{** WARNING: IEEEtran.bst: No hyphenation pattern has been}%
\typeout{** loaded for the language `#1'. Using the pattern for}%
\typeout{** the default language instead.}%
\else
\language=\csname l@#1\endcsname
\fi
#2}}
\providecommand{\BIBdecl}{\relax}
\BIBdecl

\bibitem{bonawitz2017}
K.~Bonawitz, V.~Ivanov, B.~Kreuter, A.~Marcedone, H.~B. McMahan, S.~Patel,
  D.~Ramage, A.~Segal, and K.~Seth, ``Practical secure aggregation for
  privacy-preserving machine learning,'' in \emph{proceedings of the 2017 ACM
  SIGSAC Conference on Computer and Communications Security}, 2017, pp.
  1175--1191.

\bibitem{gong2020}
M.~Gong, Y.~Xie, K.~Pan, K.~Feng, and A.~K. Qin, ``A survey on differentially
  private machine learning,'' \emph{IEEE computational intelligence magazine},
  vol.~15, no.~2, pp. 49--64, 2020.

\bibitem{mothukuri2021}
V.~Mothukuri, R.~M. Parizi, S.~Pouriyeh, Y.~Huang, A.~Dehghantanha, and
  G.~Srivastava, ``A survey on security and privacy of federated learning,''
  \emph{Future Generation Computer Systems}, vol. 115, pp. 619--640, 2021.

\bibitem{aledhari2020}
M.~Aledhari, R.~Razzak, R.~M. Parizi, and F.~Saeed, ``Federated learning: A
  survey on enabling technologies, protocols, and applications,'' \emph{IEEE
  Access}, vol.~8, pp. 140\,699--140\,725, 2020.

\bibitem{mcmahan2017}
B.~McMahan, E.~Moore, D.~Ramage, S.~Hampson, and B.~A. y~Arcas,
  ``Communication-efficient learning of deep networks from decentralized
  data,'' in \emph{Artificial intelligence and statistics}.\hskip 1em plus
  0.5em minus 0.4em\relax PMLR, 2017, pp. 1273--1282.

\bibitem{nguyen2021}
D.~C. Nguyen, M.~Ding, P.~N. Pathirana, A.~Seneviratne, J.~Li, and H.~V. Poor,
  ``Federated learning for internet of things: A comprehensive survey,''
  \emph{IEEE Communications Surveys \& Tutorials}, vol.~23, no.~3, pp.
  1622--1658, 2021.

\bibitem{zihao2022}
Z.~Zhao, M.~Luo, and W.~Ding, ``Deep leakage from model in federated
  learning,'' 2022.

\bibitem{li2021}
Q.~Li, Z.~Wen, Z.~Wu, S.~Hu, N.~Wang, Y.~Li, X.~Liu, and B.~He, ``A survey on
  federated learning systems: vision, hype and reality for data privacy and
  protection,'' \emph{IEEE Transactions on Knowledge and Data Engineering},
  2021.

\bibitem{wei2020}
K.~Wei, J.~Li, M.~Ding, C.~Ma, H.~H. Yang, F.~Farokhi, S.~Jin, T.~Q. Quek, and
  H.~V. Poor, ``Federated learning with differential privacy: Algorithms and
  performance analysis,'' \emph{IEEE Transactions on Information Forensics and
  Security}, vol.~15, pp. 3454--3469, 2020.

\bibitem{li2020}
T.~Li, A.~K. Sahu, A.~Talwalkar, and V.~Smith, ``Federated learning:
  Challenges, methods, and future directions,'' \emph{IEEE Signal Processing
  Magazine}, vol.~37, no.~3, pp. 50--60, 2020.

\bibitem{liu2020}
Y.~Liu, Y.~Kang, C.~Xing, T.~Chen, and Q.~Yang, ``A secure federated transfer
  learning framework,'' \emph{IEEE Intelligent Systems}, vol.~35, no.~4, pp.
  70--82, 2020.

\bibitem{yang2019}
Q.~Yang, Y.~Liu, T.~Chen, and Y.~Tong, ``Federated machine learning: Concept
  and applications,'' \emph{ACM Transactions on Intelligent Systems and
  Technology (TIST)}, vol.~10, no.~2, pp. 1--19, 2019.

\bibitem{gascon2016}
A.~Gasc{\'o}n, P.~Schoppmann, B.~Balle, M.~Raykova, J.~Doerner, S.~Zahur, and
  D.~Evans, ``Secure linear regression on vertically partitioned datasets.''
  \emph{IACR Cryptol. ePrint Arch.}, vol. 2016, p. 892, 2016.

\bibitem{vaidya2002}
J.~Vaidya and C.~Clifton, ``Privacy preserving association rule mining in
  vertically partitioned data,'' in \emph{Proceedings of the eighth ACM SIGKDD
  international conference on Knowledge discovery and data mining}, 2002, pp.
  639--644.

\bibitem{wu2020}
Y.~Wu, S.~Cai, X.~Xiao, G.~Chen, and B.~C. Ooi, ``Privacy preserving vertical
  federated learning for tree-based models,'' \emph{arXiv preprint
  arXiv:2008.06170}, 2020.

\bibitem{truex2019}
S.~Truex, N.~Baracaldo, A.~Anwar, T.~Steinke, H.~Ludwig, R.~Zhang, and Y.~Zhou,
  ``A hybrid approach to privacy-preserving federated learning,'' in
  \emph{Proceedings of the 12th ACM workshop on artificial intelligence and
  security}, 2019, pp. 1--11.

\bibitem{zhu2021}
H.~Zhu, H.~Zhang, and Y.~Jin, ``From federated learning to federated neural
  architecture search: a survey,'' \emph{Complex \& Intelligent Systems},
  vol.~7, no.~2, pp. 639--657, 2021.

\bibitem{jiang2022}
X.~Jiang, X.~Zhou, and J.~Grossklags, ``Comprehensive analysis of privacy
  leakage in vertical federated learning during prediction.'' \emph{Proc. Priv.
  Enhancing Technol.}, vol. 2022, no.~2, pp. 263--281, 2022.

\bibitem{Dwork2006}
C.~Dwork, K.~Kenthapadi, F.~McSherry, I.~Mironov, and M.~Naor, ``Our data,
  ourselves: Privacy via distributed noise generation,'' in \emph{Annual
  International Conference on the Theory and Applications of Cryptographic
  Techniques}.\hskip 1em plus 0.5em minus 0.4em\relax Springer, 2006, pp.
  486--503.

\bibitem{cramer2015}
R.~Cramer, I.~B. Damg{\aa}rd \emph{et~al.}, \emph{Secure multiparty
  computation}.\hskip 1em plus 0.5em minus 0.4em\relax Cambridge University
  Press, 2015.

\bibitem{sotthiwat2021}
E.~Sotthiwat, L.~Zhen, Z.~Li, and C.~Zhang, ``Partially encrypted multi-party
  computation for federated learning,'' in \emph{2021 IEEE/ACM 21st
  International Symposium on Cluster, Cloud and Internet Computing
  (CCGrid)}.\hskip 1em plus 0.5em minus 0.4em\relax IEEE, 2021, pp. 828--835.

\bibitem{madi2021}
A.~Madi, O.~Stan, A.~Mayoue, A.~Grivet-S{\'e}bert, C.~Gouy-Pailler, and
  R.~Sirdey, ``A secure federated learning framework using homomorphic
  encryption and verifiable computing,'' in \emph{2021 Reconciling Data
  Analytics, Automation, Privacy, and Security: A Big Data Challenge
  (RDAAPS)}.\hskip 1em plus 0.5em minus 0.4em\relax IEEE, 2021, pp. 1--8.

\bibitem{kanagavelu2022}
R.~Kanagavelu, Q.~Wei, Z.~Li, H.~Zhang, J.~Samsudin, Y.~Yang, R.~S.~M. Goh, and
  S.~Wang, ``Ce-fed: Communication efficient multi-party computation enabled
  federated learning,'' \emph{Array}, vol.~15, p. 100207, 2022.

\bibitem{geyer2017}
R.~C. Geyer, T.~Klein, and M.~Nabi, ``Differentially private federated
  learning: A client level perspective,'' \emph{arXiv preprint
  arXiv:1712.07557}, 2017.

\bibitem{mohammadi2021}
N.~Mohammadi, J.~Bai, Q.~Fan, Y.~Song, Y.~Yi, and L.~Liu, ``Differential
  privacy meets federated learning under communication constraints,''
  \emph{IEEE Internet of Things Journal}, pp. 1--1, 2021.

\bibitem{mcmahan2018}
H.~B. McMahan, D.~Ramage, K.~Talwar, and L.~Zhang, ``Learning differentially
  private recurrent language models,'' in \emph{International Conference on
  Learning Representations}, 2018.

\bibitem{kharitonov2019}
E.~Kharitonov, ``Federated online learning to rank with evolution strategies,''
  in \emph{Proceedings of the Twelfth ACM International Conference on Web
  Search and Data Mining}, 2019, pp. 249--257.

\bibitem{andrew2021}
G.~Andrew, O.~Thakkar, B.~McMahan, and S.~Ramaswamy, ``Differentially private
  learning with adaptive clipping,'' \emph{Advances in Neural Information
  Processing Systems}, vol.~34, pp. 17\,455--17\,466, 2021.

\bibitem{yu2021}
Z.~Yu, J.~Hu, G.~Min, Z.~Wang, W.~Miao, and S.~Li, ``Privacy-preserving
  federated deep learning for cooperative hierarchical caching in fog
  computing,'' \emph{IEEE Internet of Things Journal}, 2021.

\bibitem{liu2021}
Y.~Liu, R.~Zhao, J.~Kang, A.~Yassine, D.~Niyato, and J.~Peng, ``Towards
  communication-efficient and attack-resistant federated edge learning for
  industrial internet of things,'' \emph{ACM Transactions on Internet
  Technology (TOIT)}, vol.~22, no.~3, pp. 1--22, 2021.

\bibitem{mugunthan2019}
V.~Mugunthan, A.~Polychroniadou, D.~Byrd, and T.~H. Balch, ``Smpai: Secure
  multi-party computation for federated learning,'' in \emph{Proceedings of the
  NeurIPS 2019 Workshop on Robust AI in Financial Services}, 2019.

\bibitem{hardy2017}
S.~Hardy, W.~Henecka, H.~Ivey-Law, R.~Nock, G.~Patrini, G.~Smith, and
  B.~Thorne, ``Private federated learning on vertically partitioned data via
  entity resolution and additively homomorphic encryption,'' \emph{arXiv
  preprint arXiv:1711.10677}, 2017.

\bibitem{yang2019a}
K.~Yang, T.~Fan, T.~Chen, Y.~Shi, and Q.~Yang, ``A quasi-newton method based
  vertical federated learning framework for logistic regression,'' \emph{arXiv
  preprint arXiv:1912.00513}, 2019.

\bibitem{cheng2021}
K.~Cheng, T.~Fan, Y.~Jin, Y.~Liu, T.~Chen, D.~Papadopoulos, and Q.~Yang,
  ``Secureboost: A lossless federated learning framework,'' \emph{IEEE
  Intelligent Systems}, vol.~36, no.~6, pp. 87--98, 2021.

\bibitem{liu2020a}
Y.~Liu, Y.~Liu, Z.~Liu, Y.~Liang, C.~Meng, J.~Zhang, and Y.~Zheng, ``Federated
  forest,'' \emph{IEEE Transactions on Big Data}, 2020.

\bibitem{hou2021}
J.~Hou, M.~Su, A.~Fu, and Y.~Yu, ``Verifiable privacy-preserving scheme based
  on vertical federated random forest,'' \emph{IEEE Internet of Things
  Journal}, 2021.

\bibitem{wu13}
Y.~Wu, S.~Cai, X.~Xiao, G.~Chen, and B.~C. Ooi, ``Privacy preserving vertical
  federated learning for tree-based models,'' \emph{Proceedings of the VLDB
  Endowment}, vol.~13, no.~11.

\bibitem{feng2020}
S.~Feng and H.~Yu, ``Multi-participant multi-class vertical federated
  learning,'' \emph{arXiv preprint arXiv:2001.11154}, 2020.

\bibitem{hu2019}
Y.~Hu, D.~Niu, J.~Yang, and S.~Zhou, ``Fdml: A collaborative machine learning
  framework for distributed features,'' in \emph{Proceedings of the 25th ACM
  SIGKDD International Conference on Knowledge Discovery \& Data Mining}, 2019,
  pp. 2232--2240.

\bibitem{chen2020}
T.~Chen, X.~Jin, Y.~Sun, and W.~Yin, ``Vafl: a method of vertical asynchronous
  federated learning,'' \emph{arXiv preprint arXiv:2007.06081}, 2020.

\bibitem{wang2020}
C.~Wang, J.~Liang, M.~Huang, B.~Bai, K.~Bai, and H.~Li, ``Hybrid differentially
  private federated learning on vertically partitioned data,'' \emph{arXiv
  preprint arXiv:2009.02763}, 2020.

\bibitem{luo2021}
X.~Luo, Y.~Wu, X.~Xiao, and B.~C. Ooi, ``Feature inference attack on model
  predictions in vertical federated learning,'' in \emph{2021 IEEE 37th
  International Conference on Data Engineering (ICDE)}.\hskip 1em plus 0.5em
  minus 0.4em\relax IEEE, 2021, pp. 181--192.

\bibitem{jin2021}
X.~Jin, P.-Y. Chen, C.-Y. Hsu, C.-M. Yu, and T.~Chen, ``Cafe: Catastrophic data
  leakage in vertical federated learning,'' \emph{Advances in Neural
  Information Processing Systems}, vol.~34, pp. 994--1006, 2021.

\bibitem{weng2020}
H.~Weng, J.~Zhang, F.~Xue, T.~Wei, S.~Ji, and Z.~Zong, ``Privacy leakage of
  real-world vertical federated learning,'' \emph{arXiv preprint
  arXiv:2011.09290}, 2020.

\bibitem{sun2021}
J.~Sun, Y.~Yao, W.~Gao, J.~Xie, and C.~Wang, ``Defending against reconstruction
  attack in vertical federated learning,'' \emph{arXiv preprint
  arXiv:2107.09898}, 2021.

\bibitem{fu2022}
C.~Fu, X.~Zhang, S.~Ji, J.~Chen, J.~Wu, S.~Guo, J.~Zhou, A.~X. Liu, and
  T.~Wang, ``Label inference attacks against vertical federated learning,'' in
  \emph{31st USENIX Security Symposium (USENIX Security 22), Boston, MA}, 2022.

\bibitem{zou2022}
T.~Zou, Y.~Liu, Y.~Kang, W.~Liu, Y.~He, Z.~Yi, Q.~Yang, and Y.-Q. Zhang,
  ``Defending batch-level label inference and replacement attacks in vertical
  federated learning,'' \emph{IEEE Transactions on Big Data}, 2022.

\bibitem{dwork2014}
C.~Dwork, A.~Roth \emph{et~al.}, ``The algorithmic foundations of differential
  privacy.'' \emph{Foundations and Trends in Theoretical Computer Science},
  vol.~9, no. 3-4, pp. 211--407, 2014.

\bibitem{liu2021a}
Y.~Liu, Z.~Yi, Y.~Kang, Y.~He, W.~Liu, T.~Zou, and Q.~Yang, ``Defending label
  inference and backdoor attacks in vertical federated learning,'' \emph{arXiv
  preprint arXiv:2112.05409}, 2021.

\bibitem{abadi2016}
M.~Abadi, A.~Chu, I.~Goodfellow, H.~B. McMahan, I.~Mironov, K.~Talwar, and
  L.~Zhang, ``Deep learning with differential privacy,'' in \emph{Proceedings
  of the 2016 ACM SIGSAC conference on computer and communications security},
  2016, pp. 308--318.

\bibitem{altun2010}
K.~Altun and B.~Barshan, ``Human activity recognition using inertial/magnetic
  sensor units,'' in \emph{International workshop on human behavior
  understanding}.\hskip 1em plus 0.5em minus 0.4em\relax Springer, 2010, pp.
  38--51.

\bibitem{tsanas2012}
A.~Tsanas and A.~Xifara, ``Accurate quantitative estimation of energy
  performance of residential buildings using statistical machine learning
  tools,'' \emph{Energy and buildings}, vol.~49, pp. 560--567, 2012.

\bibitem{harrison1978}
D.~Harrison~Jr and D.~L. Rubinfeld, ``Hedonic housing prices and the demand for
  clean air,'' \emph{Journal of environmental economics and management},
  vol.~5, no.~1, pp. 81--102, 1978.

\bibitem{pace1997}
R.~K. Pace and R.~Barry, ``Sparse spatial autoregressions,'' \emph{Statistics
  \& Probability Letters}, vol.~33, no.~3, pp. 291--297, 1997.

\bibitem{abad16}
M.~Abadi, P.~Barham, J.~Chen, Z.~Chen, A.~Davis, J.~Dean, M.~Devin,
  S.~Ghemawat, G.~Irving, M.~Isard \emph{et~al.}, ``Tensorflow: a system for
  large-scale machine learning,'' in \emph{USENIX symposium on operating
  systems design and implementation (OSDI)}, 2016, pp. 265--283.

\bibitem{halimi2022federated}
A.~Halimi, S.~Kadhe, A.~Rawat, and N.~Baracaldo, ``Federated unlearning: How to
  efficiently erase a client in fl?'' \emph{arXiv preprint arXiv:2207.05521},
  2022.

\end{thebibliography}
\vfill

\end{document}